\documentclass[runningheads]{llncs}

\usepackage[T1]{fontenc}

\usepackage[dvipdfmx]{graphicx}

\usepackage{listings}
\usepackage{float}
\usepackage[dvipdfmx]{graphicx}
\usepackage{amssymb}
\usepackage{latexsym}
\usepackage{amsfonts}
\usepackage{url}
\usepackage{comment}
\usepackage{algorithm}
\usepackage{algpseudocode}

\begin{document}

\newcommand{\FIG}[3]{
\begin{minipage}[b]{#1cm}
\begin{center}
\includegraphics[width=#1cm]{#2}\\
{\scriptsize #3}
\end{center}
\end{minipage}
}

\newcommand{\FIGU}[3]{
\begin{minipage}[b]{#1cm}
\begin{center}
\includegraphics[width=#1cm,angle=180]{#2}\\
{\scriptsize #3}
\end{center}
\end{minipage}
}

\newcommand{\FIGm}[3]{
\begin{minipage}[b]{#1cm}
\begin{center}
\includegraphics[width=#1cm]{#2}\\
{\scriptsize #3}
\end{center}
\end{minipage}
}

\newcommand{\FIGR}[3]{
\begin{minipage}[b]{#1cm}
\begin{center}
\includegraphics[angle=-90,width=#1cm]{#2}
\\
{\scriptsize #3}
\vspace*{1mm}
\end{center}
\end{minipage}
}

\newcommand{\FIGRpng}[5]{
\begin{minipage}[b]{#1cm}
\begin{center}
\includegraphics[bb=0 0 #4 #5, angle=-90,clip,width=#1cm]{#2}\vspace*{1mm}
\\
{\scriptsize #3}
\vspace*{1mm}
\end{center}
\end{minipage}
}

\newcommand{\FIGpng}[5]{
\begin{minipage}[b]{#1cm}
\begin{center}
\includegraphics[bb=0 0 #4 #5, clip, width=#1cm]{#2}\vspace*{-1mm}\\
{\scriptsize #3}
\vspace*{1mm}
\end{center}
\end{minipage}
}

\newcommand{\FIGtpng}[5]{
\begin{minipage}[t]{#1cm}
\begin{center}
\includegraphics[bb=0 0 #4 #5, clip,width=#1cm]{#2}\vspace*{1mm}
\\
{\scriptsize #3}
\vspace*{1mm}
\end{center}
\end{minipage}
}

\newcommand{\FIGRt}[3]{
\begin{minipage}[t]{#1cm}
\begin{center}
\includegraphics[angle=-90,clip,width=#1cm]{#2}\vspace*{1mm}
\\
{\scriptsize #3}
\vspace*{1mm}
\end{center}
\end{minipage}
}

\newcommand{\FIGRm}[3]{
\begin{minipage}[b]{#1cm}
\begin{center}
\includegraphics[angle=-90,clip,width=#1cm]{#2}\vspace*{0mm}
\\
{\scriptsize #3}
\vspace*{1mm}
\end{center}
\end{minipage}
}

\newcommand{\FIGC}[5]{
\begin{minipage}[b]{#1cm}
\begin{center}
\includegraphics[width=#2cm,height=#3cm]{#4}~$\Longrightarrow$\vspace*{0mm}
\\
{\scriptsize #5}
\vspace*{8mm}
\end{center}
\end{minipage}
}

\newcommand{\FIGf}[3]{
\begin{minipage}[b]{#1cm}
\begin{center}
\fbox{\includegraphics[width=#1cm]{#2}}\vspace*{0.5mm}\\
{\scriptsize #3}
\end{center}
\end{minipage}
}

\newcommand{\CO}[1]{}


\newcommand{\ms}{\hspace*{10mm}}

\newcommand{\figC}{
\begin{figure*}[t]
\begin{center}
\begin{minipage}[b]{3mm}
\scriptsize
A\vspace*{1cm}\\ 
B\vspace*{1cm}\\ 
C\vspace*{1cm}\\ 
\end{minipage}\hspace*{-10mm}
\FIG{5.6}{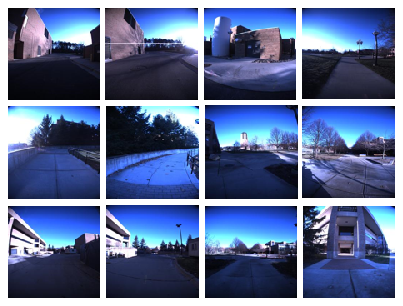}{}\FIG{5.6}{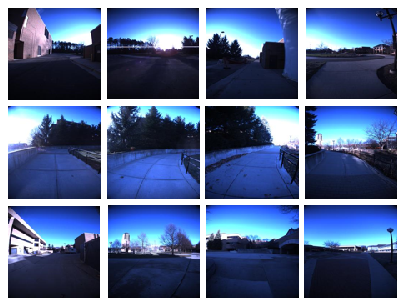}{}\\
\begin{minipage}[b]{5.7cm}
\scriptsize
\hspace*{3mm} 1st \ms 2nd \ms 3rd \ms 4th\\
\hspace*{1cm} (a) Proposed NBV planner
\end{minipage}
\begin{minipage}[b]{5.7cm}
\scriptsize
\hspace*{3mm} 1st \ms 2nd \ms 3rd \ms 4th\\
\hspace*{1cm} (b) Random planner
\end{minipage}
\caption{Example results. The view image sequences at the NBVs planned by the proposed method (left) and by the random method (right) are depicted for three different starting viewpoints. The panels from left to right show the view images at the 1st, 2nd, 3rd, and 4th viewpoints.}\label{fig:C}
\end{center}
\end{figure*}
}

\newcommand{\figD}{
\begin{figure}[t]
\begin{center}
\FIG{8}{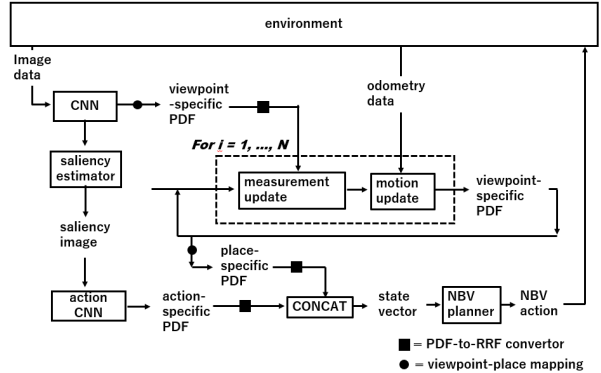}{}
\end{center}
\caption{Active VPR framework. The NBV planner is trained by transferring the state recognition ability of the CNN model for single-view VPR. Two types of cues, namely the CNN OLC (``viewpoint-specific PDV") and CNN ILC (``saliency image"), are extracted and transferred from the CNN.}\label{fig:D}
\end{figure}
}

\newcommand{\figE}{
\begin{figure}[t]
\begin{center}
\FIG{8}{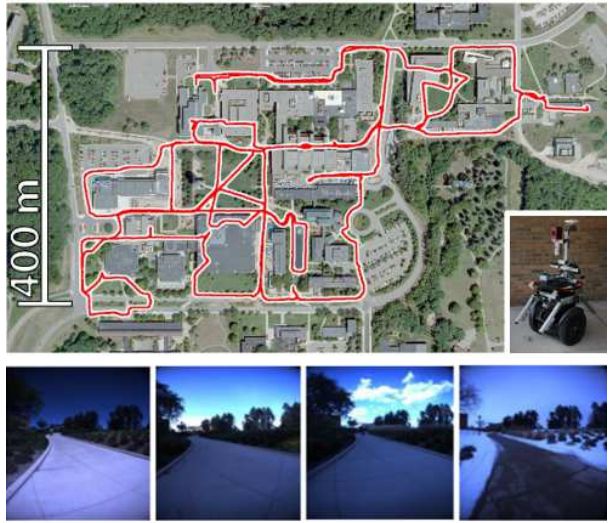}{}
\end{center}
\caption{Experimental environment. Top: entire trajectories and Segway vehicle robot. Bottom: views from onboard front-facing camera in different seasons.}\label{fig:E}
\end{figure}
}

\newcommand{\figF}{
\begin{figure}[t]
\begin{minipage}[b]{5cm}
\begin{center}
\hspace*{-20mm}\includegraphics[width=5cm]{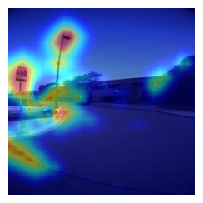}\vspace*{-1cm}\\
{\scriptsize (a) ILC in the ego-centric view.} 
\end{center}
\end{minipage}
\FIG{7}{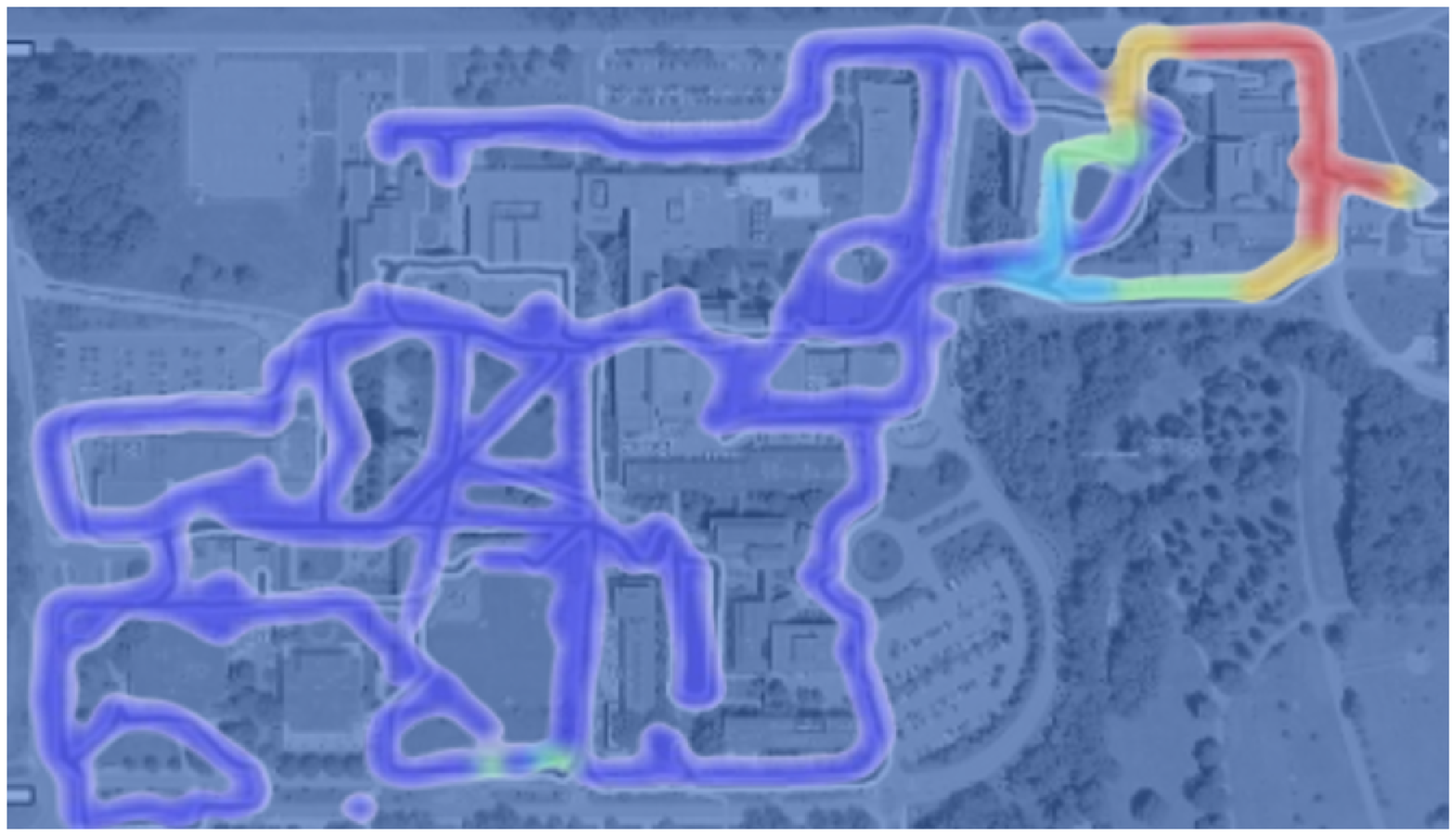}{(b) OLC in the world-centric view.} 
\caption{The training of a next-best-view (NBV) planner for visual robot place recognition (VPR) is fundamentally important for autonomous robot navigation. In this study, instead of the common path of training from visual experiences, we exploit a standard single-view VPR model using a deep CNN as the source of training data. Specifically, we divide the visual cues that are available from the CNN model into two types: the OLC and ILC, and fuse the OLC and ILC into a new state vector to reformulate the NBV planning as a domain-invariant task. Heat maps are overlaid on the images to visualize the OLC/ILC values.}\label{fig:F}
\end{figure}
}



\newcommand{\hh}{}

\newcommand{\tabA}{
\begin{table}[t]
\begin{center}
\caption{Performance results}\label{tab:A}
\begin{tabular}{r|r|r|r|r|r|}
\hline
test session & \hh{}2012/1/8 & \hh{}2012/1/15 & \hh{}2012/3/25 & \hh{}2012/8/20 & \hh{}2012/9/28 \\ \hline \hline
single-view & 0.441  & 0.293  & 0.414  & 0.345  & 0.365  \\ \hline
random & 0.547  & 0.413  & 0.538  & 0.457  & 0.542  \\ \hline
OLC-only & 0.619  & 0.471  & 0.579  & 0.497  & 0.567  \\ 
ILC-only & 0.625  & 0.457  & 0.585  & 0.494  & 0.608  \\ 
proposed & {\bf 0.647}  & {\bf 0.493}  & {\bf 0.596}  & {\bf 0.518}  & {\bf 0.623}  \\ \hline
\end{tabular}
\end{center}
\end{table}
}

\title{Active Domain-Invariant Self-Localization Using Ego-Centric and World-Centric Maps
\thanks{Our work has been supported in part by JSPS KAKENHI Grant-in-Aid
for Scientific Research (C) 17K00361, and (C) 20K12008.}}

\CO{
\author{Kanya Kurauchi\inst{1}\orcidID{0000-1111-2222-3333} \and
Kanji Tanaka\inst{1}\orcidID{1111-2222-3333-4444} \and
Ryogo Yamamoto\inst{1}\orcidID{2222--3333-4444-5555} \and
Mitsuki Yoshida\inst{1}\orcidID{2222--3333-4444-5555}}
\authorrunning{K. Kurauchi et al.}
\institute{University of Fukui, 3-9-1 bunkyo fukui, JAPAN
\email{\{mf200063, tnkknj, mf220362, mf210369\}@g.u-fukui.ac.jp}}
}

\author{Kanya Kurauchi\inst{1} \and
Kanji Tanaka\inst{1} \and
Ryogo Yamamoto\inst{1} \and
Mitsuki Yoshida\inst{1}}
\authorrunning{K. Kurauchi et al.}
\institute{University of Fukui, 3-9-1 bunkyo fukui, JAPAN
\email{\{mf200063@g., tnkknj@, mf220362@g., mf210369@g.\}u-fukui.ac.jp}}

\maketitle

\newcommand{\mykey}{Visual robot place recognition, Domain-invariant next-best-view planner, Transferring convnet features}

\begin{abstract}
The training of a next-best-view (NBV) planner for visual place recognition (VPR) is a fundamentally important task in autonomous robot navigation, for which a typical approach is the use of visual experiences that are collected in the target domain as training data. However, the collection of a wide variety of visual experiences in everyday navigation is costly and prohibitive for real-time robotic applications. We address this issue by employing a novel {\it domain-invariant} NBV planner. A standard VPR subsystem based on a convolutional neural network (CNN) is assumed to be available, and its domain-invariant state recognition ability is proposed to be transferred to train the domain-invariant NBV planner. Specifically, we divide the visual cues that are available from the CNN model into two types: the output layer cue (OLC) and intermediate layer cue (ILC). The OLC is available at the output layer of the CNN model and aims to estimate the state of the robot (e.g., the robot viewpoint) with respect to the world-centric view coordinate system. The ILC is available within the middle layers of the CNN model as a high-level description of the visual content (e.g., a saliency image) with respect to the ego-centric view. In our framework, the ILC and OLC are mapped to a state vector and subsequently used to train a multiview NBV planner via deep reinforcement learning. Experiments using the public NCLT dataset validate the effectiveness of the proposed method.
\keywords{\mykey}
\end{abstract}

\section{Introduction}

The training of a next-best-view (NBV) planner for visual place recognition (VPR) is fundamentally important for autonomous robot navigation. VPR is typically formulated as a passive single-view image classification problem in visual robot navigation \cite{classifierVPR1,classifierVPR2,classifierVPR3}, with the aim of classifying a view image into one of the predefined place classes. However, such passive formulation is ill-posed and its VPR performance is significantly dependent on the presence or absence of landmark-like objects in the view image \cite{dependency}. To address this ill-posedness, several studies have extended the passive VPR task to an active multiview VPR with NBV planning \cite{activeVPR1,activeVPR2,activeVPR3}. Active VPR is formulated as a state-to-action mapping problem that aims to maximize the VPR performance while suppressing the expected costs of action and sensing at future viewpoints. The standard approach involves training the NBV planner using the visual experience that is collected in the target domain as training data \cite{naiveTraining1,naiveTraining2,naiveTraining3}. However, this requires the collection of a large, diverse set of visual experiences for everyday navigation, which is expensive and prohibitive for real-time robotic applications.

In this study, as opposed to the common path of training from visual experiences, we exploit a deep convolutional neural network (CNN) as a source of training data (Fig. \ref{fig:F}). It has been demonstrated that a CNN classifier that is employed by a standard single-view VPR module \cite{classifierVPR1,classifierVPR2,classifierVPR3} can provide domain-invariant visual cues \cite{neuralCode1,neuralCode2,neuralCode3}, such as activations that are invoked by a view image. This motivated us to reformulate the NBV problem to use domain-invariant visual cues instead of a raw view image as the visual input to the NBV planner. Thus, synthetic visual input is used as the training data to train an NBV planner via reinforcement learning, in the spirit of the recent paradigm of simulation-based training \cite{SimBasedTrainingRL}. Because the already available CNN model is reused for supervision, it is not necessary to collect training data in the target domain.

Specifically, we divide the visual cues that are available from the CNN model into two types: the output layer cue (OLC) and intermediate layer cue (ILC). The OLC is available at the output layer of the CNN model as an estimate of the robot state (e.g., a viewpoint estimate) with respect to the world-centric view \cite{Alexnet}. 
The OLC is useful for an NBV planner to learn the optimal state-specific motion planning on a prebuilt environment map. The ILC is available within the intermediate layers of the CNN model as a high-level description of visual contents (e.g., a saliency image) with respect to the ego-centric view  \cite{Saliency}. 
The ILC is useful for an NBV planner to learn the optimal scene-specific behavior given a view image.

\figF

We subsequently exploit the OLC and ILC within a practical framework for active VPR. The ILC is implemented as a saliency image, which can be obtained within the intermediate layers of the CNN model using a saliency imaging technique \cite{Saliency}. The dimension reduction from the saliency map to a state vector is trained using a lightweight proxy NBV task. The OLC is implemented as a place-specific probabilistic distribution vector (PDV), which is obtained at the output layer of the CNN model. A sequential robot self-localization framework \cite{MarkovLocalization} is employed to integrate a sequence of place-specific PDVs from multiple viewpoints during an active VPR task. Thereafter, the OLC and ILC are fused into a compact state vector via unsupervised information fusion [18]. An NBV planner is trained as state-to-action mapping using deep Q-learning with delayed rewards [19] with the experience replay strategy. Experiments using the publicly available NCLT dataset validate the effectiveness of the proposed framework.

\section{Related Work}

Active VPR plays a particularly important role in scenarios in which robots navigate featureless environments. Typical examples of such scenarios include indoor VPR tasks in featureless passageways and classrooms, and outdoor VPR tasks in featureless road environments \cite{activeVPR1,activeVPR2,activeVPR3}. Other important scenarios include active SLAM tasks, in which an alternative VPR task known as active loop closing is considered to recognize revisits using an incomplete environmental map that is being built and often featureless \cite{activeLoopClosing1,activeLoopClosing2,activeLoopClosing3}.

Although most existing active VPR methods are non-deep, in recent years, attempts have been made to boost active VPR using deep learning.

In \cite{37}, an active VPR model known as the ``Active Neural Localizer" was trained using neural networks. This model consists of a Bayes filtering component for VPR, perceptual component for estimating the observation likelihood, and structured component for expressing beliefs. It can be trained from data end to end and achieves high VPR accuracy while minimizing the cost of action and sensing during multiview VPR tasks.

An efficient active SLAM framework was presented in \cite{40}. Although the use of learning techniques for exploration is well motivated, the end-to-end training of NBV planners is costly and prohibitive. To address this computational load, a neural SLAM module, global policy module, and local policy module were combined in a hierarchical framework. It was possible to reduce the search space during learning significantly without sacrificing the performance by means of this hierarchical module configuration.

Whereas these existing works were aimed at the efficient domain adaptation of an NBV planner, we attempt to realize a domain-invariant NBV planner that does not require adaptation to a new domain. This approach is motivated by the recent paradigm of long-term VPR, in which robots are required to be domain invariant rather than domain adaptive. However, these new learning frameworks have primarily been studied for passive VPR applications \cite{passiveLongTermVPR1,passiveLongTermVPR2,passiveLongTermVPR3}. 
Our research extends this invariant VPR framework from passive to active. We reformulate the NBV planning task as a simulation-based training problem in which the already available passive VPR module is reused as a teacher model; thus, no overhead of training costs per domain is required. Specifically, the focus of our research is on the reuse of the CNN and knowledge transfer from VPR to NBV, which has not been explored in previous studies.

\section{Approach}

Our goal is to extend a typical passive single-view VPR task to active multiview VPR. Single-view VPR is a passive image classification task that aims to predict the place class $c\in C$ from a view image $s$ for a predefined set $C$ of places. Multiview VPR aims to make a prediction from a view sequence rather than from a single view. Naturally, the VPR performance is strongly dependent on the view sequence. Hence, active VPR with viewpoint control plays an important role in multiview VPR. The training of the NBV planner is formulated as a machine-learning task with delayed rewards. That is, in the training stage, the reward for the success or failure of the multiview VPR task is delayed until the final viewpoint in each training episode. Then, in the test stage, at each step $t\in[0, T]$, the NBV action $a_t$ is planned incrementally based on the action-sensing sequence $(a_1, s_1)$$\cdots$$(a_{t-1}, s_{t-1})$, and the VPR performance at the final viewpoint $t=T$ in the test episodes is expected to be maximized.

Figure \ref{fig:D} presents the active VPR framework. 

In our approach, a CNN model is assumed to be pretrained as a visual place classifier (Section \ref{sec:cnn}), and the aim is for its domain-invariant state recognition ability to be transferred to the NBV planner. The CNN model provides two domain-invariant cues: the OLC (Section \ref{sec:bayes}) and ILC (Section \ref{sec:sal}). 

These cues are subsequently fused into a single-state vector to be transferred to the NBV planner. As no supervision is available in autonomous robotics applications, an unsupervised fusion method is adopted (Section \ref{sec:rrt}). Subsequently, an NBV planner in the form of state-to-action mapping is trained via deep Q-learning with delayed rewards (Section \ref{sec:dqn}). Each of these steps is described in detail below.

\figD

\subsection{VPR Model}\label{sec:cnn}

The CNN model is trained as a visual place classifier to classify a given view image into one of the $|C|$ predefined place classes via a standard protocol of self-supervised learning \cite{Alexnet}. The number of training epochs is set to 300,000. Prior to training, the training view sequence is partitioned into travel distances of 100 m. Accordingly, the view images along the training view sequence are divided into $|C|$  class-specific sets of training images with successive time stamps. For supervision, each training image is annotated with a pseudo-ground-truth viewpoint, which can be reconstructed from the training view sequence using a structure-from-motion technique \cite{sfm}. Notably, this process is fully self-supervised and does not require human intervention.

\subsection{CNN Output-Layer Cue}\label{sec:bayes}

A multiview VPR task aims to integrate a sequence of ego-motion and perception measurements incrementally into an estimate of the {\it viewpoint} of the robot in the form of a viewpoint-specific PDV (Fig. \ref{fig:F}b). This task is formulated as sequential robot self-localization via a Bayes filter \cite{ProbabilisticsRobotics}. It consists of two distinctive modules, namely motion updates and perception updates, as illustrated in Fig. \ref{fig:D}. The inputs to these modules are the odometry and visual measurements at each viewpoint. The output from the Bayes filter is the belief of the most recent viewpoint in the form of a viewpoint-specific PDV. The Markov localization algorithm \cite{MarkovLocalization} is employed to implement the Bayes filter. The state space is defined as a one-dimensional space that represents the travel distance along the training view sequence and a spatial resolution of 1 m is used.

The viewpoint-specific PDV that is maintained by the Bayes filter subsystem can be mutually converted into a place-specific PDV that is maintained by the VPR system. The conversion from the viewpoint-specific PDV to the place-specific PDV is defined as the operation of marginalization for each place class. The conversion from the place-specific PDV to the viewpoint-specific PDV is defined as an operation of normalization with the size of the place-specific set of viewpoints.

The $|C|$-dim place-specific PDV vector represents the knowledge to be transferred from the VPR to the NBV modules. It is also used as the final output of the multiview VPR task at the final viewpoint of a training/test episode, which is subsequently used to compute the reward in the training stage or to evaluate the performance in the test stage.

\subsection{CNN Intermediate Layer Cue}\label{sec:sal}

The aim of the saliency imaging model is to summarize where the highly complex CNN ``looks" in an image for evidence for its predictions (Fig. \ref{fig:F}a). It is trained via the saliency imaging technique in \cite{Saliency} during the VPR training process. Once it is trained, the model predicts a grayscale saliency image from the intermediate signal of the CNN. This saliency image provides pixel-wise intensity, indicating the part of the input image that is the most responsible for the decision of the classifier. Notably, the model is domain invariant; that is, it is trained only once in the training domain and no further retraining is required.

A saliency image is too high dimensional to be used directly as input into an NBV planner. Its dimensionality is proportional to the number of image pixels and has an exponential impact on the cost of simulating the possible action-sensing sequences for multiview VPR planning.

To address this issue, a dimension reduction module is trained using a proxy NBV task. The only difference between this proxy NBV task and the original NBV task is the length of the training/test episode. That is, in the proxy task, only the episodes that consist of a single action (i.e., $T=1$), which aims to classify a view image into an {\it optimal} NBV action, are considered. The optimal action for a viewpoint in a training episode is defined as the action that provides the highest VPR performance among action set $A$ at that viewpoint.

The main benefit of the use of such a proxy single-view task is that it enables the reformulation of the NBV task as an image classification task instead of the multi-view NBV planning task. That is, as opposed to the original task of multiview active VPR, reinforcement learning is not required to train such an action CNN.

Once it has been trained, the action CNN classifier can be viewed as a method for dimension reduction, which maps an input saliency image to a compact $|A|$-dim action-specific PDV. Any model architecture can be used for this dimension reduction, and in our implementation, a CNN classifier model with a standard training protocol is used.

\subsection{Reciprocal Rank Transfer}\label{sec:rrt}

A discriminative feature known as the reciprocal rank feature (RRF) is used as the input for the VPR-to-NBV knowledge transfer. In the field of multimodal information retrieval [30], the RRF is a discriminative feature that can model the output of a ranking function with unknown characteristics (e.g., a retrieval engine) and it can be used for cross-engine information fusion from multiple retrieval engines. As VPR is an instance of a ranking function, the RRF representation can be applied to an arbitrary VPR model. It should be noted that it is a ranking-based feature, and thus, it can make use of the excellent ranking ability of the CNN model.

Specifically, an RRF vector is processed in the following procedure. First, it is computed by sorting the elements in a given PDV vector in the descending order. Then, the two types of PDV cues, the OLC and ILC, are obtained in the form of $|C|$-dim and $|A|$-dim RRF vectors. Then, they are concatenated into a $(|A|+|C|)$-dim vector.

\subsection{Training NBV Planner}\label{sec:dqn}

The NBV planner is trained using a deep Q-learning network (DQN) \cite{deepQNetwork}.  
A DQN is an extension of Q-learning \cite{QLearning} that addresses the computational intractability of standard table-like value functions for high-dimensional state/action spaces. 
The basic concept of the DQN is the use of a deep neural network (instead of a table) to approximate the value function.

The RRF vector is used as the state vector for the DQN. 
The action set consists of 30 discrete action sets consisting of 1, 2, $\cdots$, and 30 m forward movements along the training viewpoint trajectory. 
The number of training episodes is set to 300,000. The experience replay technique is used to stabilize the DQN training process. 
The implementation of the DQN follows the original work in \cite{deepQNetwork}. 

The reward is determined based on whether the prediction of the multiview VPR in the training episode matches the correct answer. 
First, a sequence of odometry and image data in a training episode is integrated into an estimate of the viewpoint-specific PDV using the Bayes filter. 
The viewpoint-specific PDV is subsequently mapped onto a place-specific PDV. 
Thereafter, the place-specific PDV is translated into the top-1 place class ID. 
If this top-1 prediction result matches the ground-truth place class, a positive reward of +1 is assigned; otherwise, a negative reward of -1 is assigned.

\figE

\section{Experiments}

\subsection{Settings}

The public NCLT dataset \cite{NCLT} was used in the experiments. 
This dataset contains empirical data that were obtained by operating a Segway robot at the University of Michigan North Campus at different times of the day and in different seasons. 

Images from the front-facing camera of the onboard omnidirectional sensor LadyBug3 were used as the main modality. Moreover, GPS data from an RTK-GPS were used to reconstruct the ground truth, and to simulate the movements of the robot during the training and testing episodes.
The images exhibited various appearances, including snow cover and falling leaves, as illustrated in Fig. \ref{fig:E}. 

Notably, in our scenario of a domain-invariant NBV planner, the model was trained only once in the training domain and no further retraining for new test domains could be conducted.
Five sessions, namely
``2012/1/8," ``2012/1/15," ``2012/3/25," ``2012/8/20," and ``2012/9/28" were used as the test domains. 
The number of test episodes was set to 5,000. One session, ``2012/5/26," 
which had one of the largest area coverages and longest travel distances (6.3 km), was used as the training domain.

The number of actions per episode was set to $T=3$. 
That is, the estimate from the third viewpoint of the Bayes filter, 
which integrated measurements from the 0-th $\cdots$ 3rd viewpoints, was used as the final output of our active VPR system.

The dataset was preprocessed as follows: The viewpoint trajectories of the robot were discretized using a spatial resolution of 1 m in terms of the travel distance. When multiple view images belonged to the same discretized viewpoint, the image with the youngest time stamp was used. The starting point of each episode was randomly selected. For an episode in which the starting viewpoint was very close to the unseen area in the workspace, the robot was often forced out by an action to the unseen area, and such a test episode was simply discarded and replaced with a newly sampled episode.

\subsection{Comparative Methods}

The proposed method was compared with four comparative methods: the single-view method, random method, OLC-only method, and ILC-only method.
A passive single-view VPR scenario was assumed in the single-view method. This method could be viewed as a baseline for verifying the significance of multiview methods.
The random method was a naive multiview method in which the action at each viewpoint in each episode was randomly sampled from action set $A$. 
This method could be viewed as a baseline for verifying the significance of the NBV planning.
The OLC-only method was an ablation of the proposed method in which the NBV planner used the OLC as a $|C|$-dim state vector.
The ILC-only method was another ablation of the proposed method in which the NBV planner used the ILC as an $|A|$-dim state vector.

\subsection{Performance Index}

The mean reciprocal rank (MRR) \cite{MRR} was used as the performance index. 
The MRR is commonly used to evaluate information retrieval algorithms, where a larger MRR indicates better performance. 
The maximum value of MRR = 1 when the ground-truth class can be answered first for all queries (i.e., test episodes). 
The minimum value of MRR = 0 when the correct answer does not exist within the shortlist for all queries.

\subsection{Results}

The proposed method and four comparative methods were evaluated. 
We were particularly interested in investigating the robustness of the viewpoint planning against changes in the environments 
and the contribution of the viewpoint planning to improving the recognition performance.

Figure \ref{fig:C} depicts examples of view sequences along the planned viewpoint trajectories for the proposed and random methods. 
The results of these two methods exhibited a clear contrast. 
For example A, 
the proposed NBV method yielded a good viewpoint at which landmark-like objects such as tubular buildings and streets were observed, 
whereas the random method yielded featureless scenes. 
For example B, major strategic forward movements were planned by the proposed method to avoid featureless scenes, whereas short travel distance movements by which the robot remained within featureless areas were yielded by the random method. 
For example C, the proposed method was successful in finally moving to the front of the characteristic building, whereas the random method provided many observations that did not include landmark-like objects.

\tabA
\figC

Table \ref{tab:A} presents the performance results. 
It can be observed that the performance of the single-view method deteriorated significantly in the test on ``2012/1/15". 
This is because the overall appearance of the scene was very different from that in the training domain, mainly owing to snow cover. 
A comparison of the random and single-view methods revealed that the former outperformed the latter in all test data, 
according to which the significance of the multiview VPR was confirmed. 
The ILC-only method outperformed the random method in all test data, 
indicating that the ILC is effective for active VPR. 
A comparison of the ILC-only method and OLC-only method demonstrated that the superiority or inferiority of these two methods is significantly dependent on the type of test scene. 
As expected, the OLC was relatively ineffective from the early viewpoints of the multiview VPR task, as inferences from the measurement history could not be leveraged. 
However, the ILC was relatively ineffective when the input scene was not discriminative because the saliency image was less reliable. 
It can be concluded that these two methods have complementary roles with different advantages and disadvantages. 
Importantly, the proposed method, which combines the advantages of the two methods, outperformed the individual methods in all test domains.

\section{Conclusions}

A new training method for NBV planners in VPR has been presented. In the proposed approach, simulation-based training is realized by employing the available CNN model as a teacher model. 
Specifically, a novel task of transferring ConvNet features from passive to active robot self-localization is addressed and the use of ego-centric and world-centric views is proposed.
The domain-invariant state recognition capability of the CNN model is transferred and the NBV planner is trained to be domain invariant. Furthermore, the two types of independent visual cues have been proposed for extraction from this CNN model. Experiments using the public NCLT dataset demonstrated the effectiveness of the proposed method.

\bibliography{kurauchi} 

\begin{thebibliography}{10}

\bibitem{classifierVPR1}
Yiyi Liao, Sarath Kodagoda, Yue Wang, Lei Shi, and Yong Liu.
\newblock Place classification with a graph regularized deep neural network.
\newblock {\em {IEEE} Trans. Cogn. Dev. Syst.}, 9(4):304--315, 2017.

\bibitem{classifierVPR2}
Andry Chowanda and Rhio Sutoyo.
\newblock Deep learning for visual indonesian place classification with
  convolutional neural networks.
\newblock {\em Procedia Computer Science}, 157:436--443, 2019.

\bibitem{classifierVPR3}
Giseop Kim, Byungjae Park, and Ayoung Kim.
\newblock 1-day learning, 1-year localization: Long-term lidar localization
  using scan context image.
\newblock {\em {IEEE} Robotics Autom. Lett.}, 4(2):1948--1955, 2019.

\bibitem{dependency}
Gabriele~Moreno Berton, Carlo Masone, Valerio Paolicelli, and Barbara Caputo.
\newblock Viewpoint invariant dense matching for visual geolocalization.
\newblock In {\em 2021 {IEEE/CVF} International Conference on Computer Vision,
  {ICCV} 2021, Montreal, QC, Canada, October 10-17, 2021}, pages 12149--12158.
  {IEEE}, 2021.

\bibitem{activeVPR1}
Emanuele Frontoni and Primo Zingaretti.
\newblock A vision based algorithm for active robot localization.
\newblock In {\em {CIRA} 2005, International Symposium on Computational
  Intelligence in Robotics and Automation, June 27-30, 2005, Espoo, Finland},
  pages 347--352. {IEEE}, 2005.

\bibitem{activeVPR2}
Koosha Khalvati and Alan~K. Mackworth.
\newblock Active robot localization with macro actions.
\newblock In {\em 2012 {IEEE/RSJ} International Conference on Intelligent
  Robots and Systems, {IROS} 2012, Vilamoura, Algarve, Portugal, October 7-12,
  2012}, pages 187--193. {IEEE}, 2012.

\bibitem{activeVPR3}
Gian~Luca Mariottini and Stergios~I. Roumeliotis.
\newblock Active vision-based robot localization and navigation in a visual
  memory.
\newblock In {\em {IEEE} International Conference on Robotics and Automation,
  {ICRA} 2011, Shanghai, China, 9-13 May 2011}, pages 6192--6198. {IEEE}, 2011.

\bibitem{naiveTraining1}
Yuke Zhu, Roozbeh Mottaghi, Eric Kolve, Joseph~J. Lim, Abhinav Gupta,
  Li~Fei{-}Fei, and Ali Farhadi.
\newblock Target-driven visual navigation in indoor scenes using deep
  reinforcement learning.
\newblock In {\em 2017 {IEEE} International Conference on Robotics and
  Automation, {ICRA} 2017, Singapore, Singapore, May 29 - June 3, 2017}, pages
  3357--3364. {IEEE}, 2017.

\bibitem{naiveTraining2}
Hee~Rak Beom and Hyung~Suck Cho.
\newblock A sensor-based navigation for a mobile robot using fuzzy logic and
  reinforcement learning.
\newblock {\em {IEEE} Trans. Syst. Man Cybern.}, 25(3):464--477, 1995.

\bibitem{naiveTraining3}
Henrik Kretzschmar, Markus Spies, Christoph Sprunk, and Wolfram Burgard.
\newblock Socially compliant mobile robot navigation via inverse reinforcement
  learning.
\newblock {\em Int. J. Robotics Res.}, 35(11):1289--1307, 2016.

\bibitem{neuralCode1}
Abin Jose, Timo Horstmann, and Jens{-}Rainer Ohm.
\newblock Optimized binary hashing codes generated by siamese neural networks
  for image retrieval.
\newblock In {\em 26th European Signal Processing Conference, {EUSIPCO} 2018,
  Roma, Italy, September 3-7, 2018}, pages 1487--1491. {IEEE}, 2018.

\bibitem{neuralCode2}
Stephen Hausler, Sourav Garg, Ming Xu, Michael Milford, and Tobias Fischer.
\newblock Patch-netvlad: Multi-scale fusion of locally-global descriptors for
  place recognition.
\newblock In {\em {IEEE} Conference on Computer Vision and Pattern Recognition,
  {CVPR} 2021, virtual, June 19-25, 2021}, pages 14141--14152. Computer Vision
  Foundation / {IEEE}, 2021.

\bibitem{neuralCode3}
Yangyang Wang, Xiaorui Ma, Jie Wang, Shilong Hou, Ju~Dai, Dongbing Gu, and
  Hongyu Wang.
\newblock Robust auv visual loop closure detection based on variational
  auto-encoder network.
\newblock {\em IEEE Transactions on Industrial Informatics}, pages 1--1, 2022.

\bibitem{SimBasedTrainingRL}
Wenshuai Zhao, Jorge~Pe{\~{n}}a Queralta, and Tomi Westerlund.
\newblock Sim-to-real transfer in deep reinforcement learning for robotics: a
  survey.
\newblock In {\em 2020 {IEEE} Symposium Series on Computational Intelligence,
  {SSCI} 2020, Canberra, Australia, December 1-4, 2020}, pages 737--744.
  {IEEE}, 2020.

\bibitem{Alexnet}
Alex Krizhevsky, Ilya Sutskever, and Geoffrey~E. Hinton.
\newblock Imagenet classification with deep convolutional neural networks.
\newblock {\em Commun. {ACM}}, 60(6):84--90, 2017.

\bibitem{Saliency}
Ruth~C. Fong and Andrea Vedaldi.
\newblock Interpretable explanations of black boxes by meaningful perturbation.
\newblock In {\em {IEEE} International Conference on Computer Vision, {ICCV}
  2017, Venice, Italy, October 22-29, 2017}, pages 3449--3457. {IEEE} Computer
  Society, 2017.

\bibitem{MarkovLocalization}
Wolfram Burgard, Dieter Fox, and Sebastian Thrun.
\newblock Markov localization for mobile robots in dynamic environments.
\newblock {\em CoRR}, abs/1106.0222, 2011.

\bibitem{activeLoopClosing1}
Cyrill Stachniss, Dirk H{\"{a}}hnel, and Wolfram Burgard.
\newblock Exploration with active loop-closing for fastslam.
\newblock In {\em 2004 {IEEE/RSJ} International Conference on Intelligent
  Robots and Systems, Sendai, Japan, September 28 - October 2, 2004}, pages
  1505--1510. {IEEE}, 2004.

\bibitem{activeLoopClosing2}
Hannah Lehner, Martin~J. Schuster, Tim Bodenm{\"{u}}ller, and Simon Kriegel.
\newblock Exploration with active loop closing: {A} trade-off between
  exploration efficiency and map quality.
\newblock In {\em 2017 {IEEE/RSJ} International Conference on Intelligent
  Robots and Systems, {IROS} 2017, Vancouver, BC, Canada, September 24-28,
  2017}, pages 6191--6198. {IEEE}, 2017.

\bibitem{activeLoopClosing3}
Eungchang~Mason Lee, Junho Choi, Hyungtae Lim, and Hyun Myung.
\newblock {REAL:} rapid exploration with active loop-closing toward large-scale
  3d mapping using uavs.
\newblock In {\em {IEEE/RSJ} International Conference on Intelligent Robots and
  Systems, {IROS} 2021, Prague, Czech Republic, September 27 - Oct. 1, 2021},
  pages 4194--4198. {IEEE}, 2021.

\bibitem{37}
Devendra~Singh Chaplot, Emilio Parisotto, and Ruslan Salakhutdinov.
\newblock Active neural localization.
\newblock {\em arXiv preprint arXiv:1801.08214}, 2018.

\bibitem{40}
Devendra~Singh Chaplot, Dhiraj Gandhi, Saurabh Gupta, Abhinav Gupta, and Ruslan
  Salakhutdinov.
\newblock Learning to explore using active neural {SLAM}.
\newblock In {\em 8th International Conference on Learning Representations,
  {ICLR} 2020, Addis Ababa, Ethiopia, April 26-30, 2020}. OpenReview.net, 2020.

\bibitem{passiveLongTermVPR1}
Zetao Chen, Lingqiao Liu, Inkyu Sa, Zongyuan Ge, and Margarita Chli.
\newblock Learning context flexible attention model for long-term visual place
  recognition.
\newblock {\em {IEEE} Robotics Autom. Lett.}, 3(4):4015--4022, 2018.

\bibitem{passiveLongTermVPR2}
Fei Han, Xue Yang, Yiming Deng, Mark Rentschler, Dejun Yang, and Hao Zhang.
\newblock {SRAL:} shared representative appearance learning for long-term
  visual place recognition.
\newblock {\em {IEEE} Robotics Autom. Lett.}, 2(2):1172--1179, 2017.

\bibitem{passiveLongTermVPR3}
Diwei Sheng, Yuxiang Chai, Xinru Li, Chen Feng, Jianzhe Lin, Cl{\'{a}}udio~T.
  Silva, and John{-}Ross Rizzo.
\newblock {NYU-VPR:} long-term visual place recognition benchmark with view
  direction and data anonymization influences.
\newblock In {\em {IEEE/RSJ} International Conference on Intelligent Robots and
  Systems, {IROS} 2021, Prague, Czech Republic, September 27 - Oct. 1, 2021},
  pages 9773--9779. {IEEE}, 2021.

\bibitem{sfm}
Johannes~L Schonberger and Jan-Michael Frahm.
\newblock Structure-from-motion revisited.
\newblock In {\em Proceedings of the IEEE conference on computer vision and
  pattern recognition}, pages 4104--4113, 2016.

\bibitem{ProbabilisticsRobotics}
Sebastian Thrun, Wolfram Burgard, and Dieter Fox.
\newblock {\em Probabilistic robotics}.
\newblock Intelligent robotics and autonomous agents. {MIT} Press, 2005.

\bibitem{deepQNetwork}
Volodymyr Mnih, Koray Kavukcuoglu, David Silver, Andrei~A. Rusu, Joel Veness,
  Marc~G. Bellemare, Alex Graves, Martin~A. Riedmiller, Andreas Fidjeland,
  Georg Ostrovski, Stig Petersen, Charles Beattie, Amir Sadik, Ioannis
  Antonoglou, Helen King, Dharshan Kumaran, Daan Wierstra, Shane Legg, and
  Demis Hassabis.
\newblock Human-level control through deep reinforcement learning.
\newblock {\em Nat.}, 518(7540):529--533, 2015.

\bibitem{QLearning}
S{\'{e}}rgio~F. Chevtchenko and Teresa~Bernarda Ludermir.
\newblock Learning from sparse and delayed rewards with a multilayer spiking
  neural network.
\newblock In {\em 2020 International Joint Conference on Neural Networks,
  {IJCNN} 2020, Glasgow, United Kingdom, July 19-24, 2020}, pages 1--8. {IEEE},
  2020.

\bibitem{NCLT}
Nicholas Carlevaris{-}Bianco, Arash~K. Ushani, and Ryan~M. Eustice.
\newblock University of michigan north campus long-term vision and lidar
  dataset.
\newblock {\em Int. J. Robotics Res.}, 35(9):1023--1035, 2016.

\bibitem{MRR}
Nick Craswell.
\newblock Mean reciprocal rank.
\newblock {\em Encyclopedia of database systems}, 1703, 2009.

\end{thebibliography}
\bibliographystyle{unsrt} 

\end{document}